\documentclass[journal]{IEEEtran}
\usepackage[utf8]{inputenc}
\usepackage[T1]{fontenc}% optional T1 font encoding
\usepackage{textgreek}
\usepackage{graphicx}
\usepackage{times}
\usepackage{helvet}
\usepackage{courier}
\usepackage{amsmath}
\usepackage{algorithm}
\usepackage{algorithmic}
\usepackage{csquotes} 
\usepackage{color}
\usepackage{paralist}
\usepackage{amssymb}
\usepackage{indentfirst}
\usepackage{subfigure}
\usepackage{float}
\usepackage{multirow}
\usepackage{cite}
\usepackage{mathrsfs}

\usepackage{mathrsfs} 
\usepackage{amsfonts}
\usepackage{todonotes}
\usepackage{pgfplots} %引用包
\usepackage{epstopdf}
\usepackage{epsfig}
\pgfplotsset{compat=newest}
\usepackage{color}
\definecolor{forestgreen}{RGB}{0,139,69}

\usepackage{xcolor}
\definecolor{citecolor}{HTML}{0071bc}
\usepackage[colorlinks, linkcolor=red,  anchorcolor=blue, citecolor=citecolor]{hyperref} 

\definecolor{SeaGreen4}{RGB}{0,205,102} 
\definecolor{SlateBlue}{RGB}{106,90,205} 
\definecolor{DarkRed}{RGB}{178,34,34} 
	
\usepackage[switch]{lineno}
\usepackage{textcomp,booktabs}
\usepackage{amssymb}% http://ctan.org/pkg/amssymb
\usepackage{pifont}% http://ctan.org/pkg/pifont
\newcommand{\cmark}{\ding{51}}%
\usepackage{makecell}

\usepackage{colortbl}
\definecolor{mygray}{gray}{.9}
\definecolor{mypink}{rgb}{.99,.91,.95}
\definecolor{mycyan}{cmyk}{.3,0,0,0}

\begin{document}

\title{ Mamba-FETrack V2: Revisiting State Space Model for Frame-Event based Visual Object Tracking}

\author{Shiao Wang, Ju Huang, Qingchuan Ma, Jinfeng Gao, Chunyi Xu, Xiao Wang*, \emph{Member, IEEE}, \\ 
    Lan Chen*, Bo Jiang 

\thanks{$\bullet$ Shiao Wang, Ju Huang, Qingchuan Ma, Jinfeng Gao, Chunyi Xu, Xiao Wang, and Bo Jiang are all with the School of Computer Science and Technology, Anhui University, Hefei 230601, China. (email: wsa1943230570@126.com, huangju991011@163.com, \{xiaowang, jiangbo\}@ahu.edu.cn, 1467662818@qq.com, mqc20021030@gmail.com, chunyi871@gmail.com)} 

\thanks{$\bullet$ Lan Chen is with the School of Electronic and Information Engineering, Anhui University, Hefei 230601, China. (email: chenlan@ahu.edu.cn)}

\thanks{* Corresponding Author: Xiao Wang \& Lan Chen} 
}

\markboth{ IEEE Transactions on ***, 2025 } 
{Shell \MakeLowercase{\textit{et al.}}: Bare Demo of IEEEtran.cls for IEEE Journals}

% make the title area
\maketitle

% As a general rule, do not put math, special symbols or citations in the abstract or keywords.
\begin{abstract}
Combining traditional RGB cameras with bio-inspired event cameras for robust object tracking has garnered increasing attention in recent years. However, most existing multimodal tracking algorithms depend heavily on high-complexity Vision Transformer architectures for feature extraction and fusion across modalities. This not only leads to substantial computational overhead but also limits the effectiveness of cross-modal interactions.
In this paper, we propose an efficient RGB-Event object tracking framework based on the linear-complexity Vision Mamba network, termed Mamba-FETrack V2. Specifically, we first design a lightweight Prompt Generator that utilizes embedded features from each modality, together with a shared prompt pool, to dynamically generate modality-specific learnable prompt vectors. These prompts, along with the modality-specific embedded features, are then fed into a Vision Mamba-based FEMamba backbone, which facilitates prompt-guided feature extraction, cross-modal interaction, and fusion in a unified manner. Finally, the fused representations are passed to the tracking head for accurate target localization.
Extensive experimental evaluations on multiple RGB-Event tracking benchmarks, including short-term COESOT dataset and long-term datasets, i.e., FE108 and FELT V2, demonstrate the superior performance and efficiency of the proposed tracking framework.
The source code and pre-trained models will be released on \url{https://github.com/Event-AHU/Mamba_FETrack}. 
\end{abstract}


\IEEEpeerreviewmaketitle

\section{Introduction}

%% background 
\IEEEPARstart{V}{isual} Object Tracking (VOT) is a crucial task in the field of computer vision, aiming to locate a given target in subsequent video frames, given its initial position in the first frame. This task demonstrates significant practical value, covering a wide range of important fields such as security surveillance, autonomous driving perception, sports analytics, and human-computer interaction. Currently, most visual object tracking algorithms~\cite{zhang2024augment, chen2025improving, wu2025local, xun2024linker} are designed and developed based on RGB cameras. With the continuous advancement of deep learning technologies, these unimodal tracking approaches have demonstrated exceptional performance in specific scenarios.
Despite the good performance achieved, RGB cameras still have certain limitations. In conditions such as low light, overexposure, and fast motion, the data captured by RGB cameras often performs poorly, affecting tracking accuracy.

To overcome the inherent limitations of conventional RGB cameras, researchers have begun to combine other auxiliary modalities to address challenges under extreme conditions, leading to the emergence of bio-inspired event cameras (e.g., DVS 346, Prophesee, CeleX-V). Unlike the RGB cameras, event cameras perceive information by capturing changes in the light intensity of a scene. When the light intensity changes exceed a given threshold, the event camera generates an event signal at the corresponding pixel location. This event signal is represented by a quadruple $\{x, y, t, p\}$, where $(x, y)$ denotes the spatial location of the event, $t$ represents the timestamp of the event, and $p$ indicates the polarity of the light intensity change. Due to the unique data acquisition principle of event cameras, they offer advantages such as high dynamic range, high temporal resolution, and low power consumption. Consequently, event cameras often demonstrate significant advantages over frame-based RGB cameras, particularly in challenging environments where the latter struggle.

\begin{figure*}
\centering
\includegraphics[width=\textwidth]{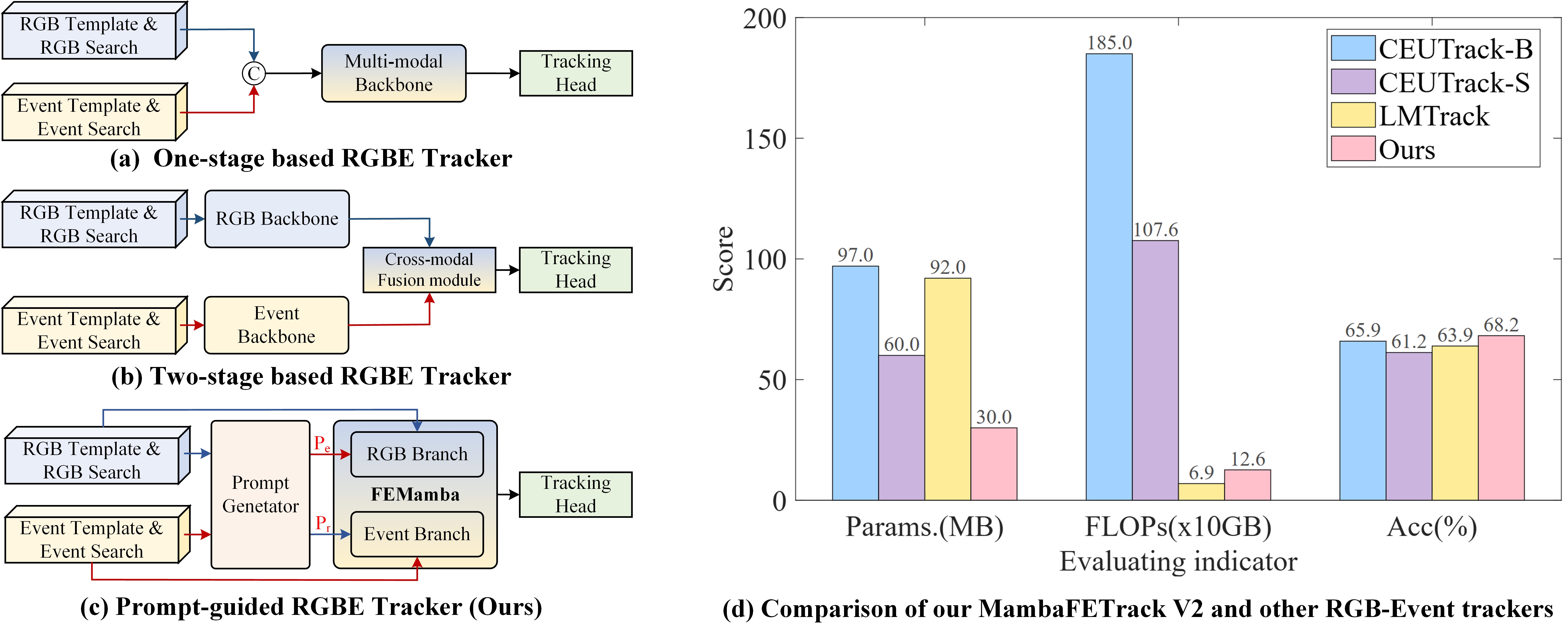}
\caption{Architectural comparison of (a) one-stage based and (b) two-stage based trackers, as well as (c) our proposed prompt-guided RGB-Event tracker. (d) A comparison between our proposed Mamba-FETrack V2 method and other robust RGB-Event tracking models in terms of parameters, FLOPs, and accuracy.}
\label{fig:First_Img} 
\end{figure*}

Although event cameras have the advantages mentioned above, they also have certain limitations, such as an inability to capture fine texture details of targets. Fortunately, this is exactly where RGB cameras excel. Therefore, the idea of combining the strengths of both event cameras and RGB cameras for collaborative visual object tracking has gradually gained increasing attention from researchers, as shown in Fig.~\ref{fig:First_Img} (a) and (b). For example, Tang et al.~\cite{tang2022coesot} propose a one-stage network architecture for unified tracking, which simultaneously achieves feature extraction, fusion, matching, and interactive learning. Zhang et al.~\cite{Zhang2023FrameEventAA} propose a framework for high frame rate tracking-event alignment and fusion network, which significantly improves the performance of high frame rate tracking by combining the advantages of traditional RGB cameras and event cameras.

The efficacy of the aforementioned multimodal fusion methods heavily relies on the backbone networks' capability to process heterogeneous visual data. Common algorithms include those Convolutional Neural Networks~\cite{lecun1998gradient} (CNN) based (e.g., ATOM~\cite{danelljan2019atom}, DiMP~\cite{bhat2019learning}, Ocean~\cite{zhang2020ocean}), and Vision Transformer~\cite{dosovitskiy2020image} (ViT) based (e.g., STARK~\cite{yan2021learning}, AiATrack~\cite{gao2022aiatrack}, SwinTrack~\cite{lin2022swintrack}), each with its own characteristics and playing a key role in various tasks. However, the limited local modeling capability of CNNs and the high computational complexity of ViTs fundamentally constrain their performance across various application scenarios. This has motivated recent exploration of State Space Models (SSMs)~\cite{gu2022efficiently}, particularly Vision Mamba architectures~\cite{vim, liu2024vmamba, hatamizadeh2025mambavision}, which offer linear computational scaling while maintaining global receptive fields.

In this work, we propose a novel Vision Mamba-based RGB-Event visual object tracking method, termed Mamba-FETrack V2, which aims to leverage the Mamba network's exceptional modeling capacity and linear computational complexity to achieve more efficient and accurate visual object tracking. Specifically, as shown in Fig.~\ref{fig:First_Img} (c), inspired by prompt learning~\cite{liu2023pre}, we design a novel Prompt Generator that encodes the information from RGB and event modalities to generate two modality-specific, learnable prompts, effectively capturing the unique characteristics of each modality. Subsequently, we introduce a unified and more compact Vision Mamba-based backbone network designed to simultaneously perform cross-modal feature extraction, fusion, and interaction. To enable effective interaction between modalities, inspired by MambaIRv2~\cite{guo2025mambairv2}, we introduce an innovative state space integration strategy, in which the prompt of each modality is additively fused with the state matrix of the counterpart modality within the SSM module. This design facilitates efficient cross-modal information propagation through learned prompt representations. As shown in Fig.~\ref{fig:First_Img} (d), our proposed method achieves a good balance between model complexity and accuracy.

% and State Space Models (e.g., Mamba-FETrack~\cite{huang2024mamba}, TrackingMamba~\cite{wang2024trackingmamba}, TemTrack~\cite{xie2025robust}), 

To sum up, the main contributions of this paper can be summarized into the following three aspects: 

1). We propose a unified Vision Mamba-based framework, Mamba-FETrack V2, which is designed to jointly model RGB and event-based modalities for effective object tracking.

2). We introduce a novel Prompt Generator for RGB-Event modality fusion and establish effective cross-modal interaction through a prompt-guided fusion strategy.

3). We demonstrate the effectiveness and efficiency of our proposed method through extensive experiments on datasets such as FE108, FELT V2, and COESOT.

This paper is an extension of our Mamba-FETrack~\cite{huang2024mamba}, which was published on Pattern Recognition and Computer Vision (PRCV) 2024. The main changes of this paper can be summarized as follows: 
\textbf{(I). New feature modeling framework:} We systematically enhanced the framework by replacing the two-stage feature modeling with a unified architecture that seamlessly integrates feature extraction, fusion, and interaction.
\textbf{(II). More efficient feature fusion and interaction strategy:} We employ a Prompt Generator to produce modality-specific prompts and introduce an effective prompt-guided cross-fusion mechanism to replace the Mamba fusion module in the previous work, thereby improving the efficiency of cross-modal feature fusion and interaction.
\textbf{(III). More comprehensive experiments:} We conduct a more detailed and comprehensive set of experiments to demonstrate the effectiveness of the proposed method. Building upon the original framework, our Mamba-FETrack V2 achieves significant improvements in both efficiency and robustness for object tracking.

\textit{The remainder of this paper is organized as follows:} 
The related works are provided in Section~\ref{sec::relatedWorks}, focusing on RGB-Event based Tracking, Mamba Network, and Prompt-based Learning. 
For the proposed approach, we first give an overview in Section~\ref{subsec:overview}, then dive into the details of input representation, Prompt Generator module, FEMamba backbone network, tracking head, and loss function in subsequent sub-sections. 
In Section~\ref{sec::Experiments}, we conduct the experiments on multiple benchmark datasets and both the qualitative and quantitative experiments fully validate the effectiveness of our proposed Mamba-FETrack V2 framework for RGB-Event based visual object tracking. 
Finally, we conclude this paper and propose possible research directions for the Mamba-based visual object tracking in Section~\ref{sec::conclusion}.

\section{Related Works} \label{sec::relatedWorks}

In this section, we mainly introduce the related works on the RGB-Event based Tracking, Mamba Network, and Prompt-based Learning. More related works can be found in the surveys~\cite{zhang2024awesome, zhang2024survey, lei2024prompt, wang2024state} and the paper list\footnote{\url{github.com/wangxiao5791509/Single_Object_Tracking_Paper_List}}. 

\subsection{RGB-Event based Tracking} 
In recent years, event cameras have been receiving increasing attention, and multimodal object tracking algorithms that integrate RGB cameras and event cameras have demonstrated strong performance. 
Tang et al.~\cite{tang2022coesot} propose CEUTrack, which models the relationship between RGB frames and event voxels using a unified Transformer network. 
Zhang et al.~\cite{zhang2021object} propose a multimodal tracking approach that combines frame and event domains and balances the contributions of both domains.
Wang et al.~\cite{Wang2021VisEventRO} propose a reliable object tracking method through the collaboration of two different sensors (RGB camera and event camera). 
TENet~\cite{shao2025tenet} designs a targetness entanglement strategy to promote the selective enhancement of object-related cues in both modalities. 
Liu et al.~\cite{liu2024emtrack} propose EMTrack, achieving real-time speed on limited resource devices while maintaining performance using multimodal experts.
SDSTrack~\cite{hou2024sdstrack} introduces a self-distillation symmetric adapter framework, achieving effective fusion between RGB and event data. 
Hu et al.~\cite{hu2025exploiting} propose STTrack, which effectively improves tracking performance in complex scenes by introducing a spatiotemporal state generator.
OneTracker~\cite{hong2024onetracker} unifies RGB-Event tracking within a basic model framework, using efficient fine-tuning and visual language prompt schemes.
Zhu et al.~\cite{Zhu2023CrossmodalOH} propose a mask modeling strategy that randomly masks tokens across modalities to encourage the Vision Transformer (ViT) to close the distribution gap between them, thus improving the model's capability.
Despite these advances, the high computational complexity of Transformer networks imposes a heavy computational burden on tracking algorithms. Unlike the aforementioned methods, we are inspired by the linear complexity advantage of the Mamba model and propose a Vision Mamba-based RGB-Eventobject tracking algorithm.

\subsection{Mamba Network}
% Recent advances in sequence modeling have highlighted the limitations of traditional RNNs due to their instability during training, which motivated early works like Unitary Evolution RNNs ~\cite{arjovsky2016unitary} to constrain the eigenvalues of transition matrices, thereby ensuring stable gradients. Building upon this direction, the Mamba~\cite{Gu2023MambaLS} architecture was introduced as a linear-time sequence modeling framework based on the Selective State Space Model, capable of handling long-context data efficiently across domains such as vision, language, audio, and genomics.

The State Space Model (SSM) was originally developed as a mathematical framework to describe the dynamics of time-dependent systems. To enhance its modeling capacity and computational efficiency, Gu et al.\cite{Gu2021CombiningRC} proposed the Linear State Space Layer (LSSL), which integrates the strengths of RNNs, temporal convolutions, and neural differential equations (NDEs). To overcome the gradient vanishing/exploding issues in long-sequence modeling, the HiPPO framework\cite{gu2020hippo} introduces recursive memory and optimal polynomial projection, enabling more effective handling of long-term dependencies. Building on these advances, Gu et al.~\cite{gu2022efficiently} proposed the Structured State Space Sequence Model (S4), a novel parameterization of SSMs that captures long-range dependencies while significantly improving computational efficiency through both theoretical and empirical innovations.

Recently, State Space Models (SSMs), initially designed for natural language processing tasks, have gained increasing attention. Mamba~\cite{Gu2023MambaLS} introduced a time-varying SSM equipped with a selective mechanism, demonstrating strong capability in modeling long sequences. Its success has inspired researchers to explore its application across a broad range of fields.
In the field of computer vision, Vision Mamba~\cite{vim} applies a bidirectional SSM design to the field of vision, achieving competitive results in tasks such as image classification, object detection, and semantic segmentation.
VMamba~\cite{liu2024vmamba} introduces the Cross-Scan Module (CSM) to traverse spatial dimensions and reorganize non-causal visual inputs into ordered patch sequences.
Hatamizadeh et al.~\cite{hatamizadeh2025mambavision} propose MambaVision, a hybrid visual backbone network that combines Mamba and Transformer structures, enhancing its ability to efficiently model visual features. 
In reinforcement learning, Decision Mamba~\cite{lv2024decision} incorporates self-evolution regularization to improve performance on noisy offline trajectories, surpassing existing baselines. 
For multimodal learning, Coupled Mamba~\cite{li2024coupledmamba} proposes a coupled SSM that enhances both inter- and intra-modal correlation modeling, showing strong fusion capabilities.
Xie et al.~\cite{xie2024fusionmamba} propose a novel dynamic feature enhancement framework called FusionMamba for multimodal image fusion tasks.
In this work, we take advantage of the Mamba network’s linear computational complexity to achieve more efficient visual object tracking.

% Further developments include DiMSUM~\cite{zhou2024dimsum}, a diffusion-based model that unifies spatial and frequency domains for image generation. MambaLRP~\cite{yang2024mambalrp}, which improves model interpretability through stable layer-wise relevance propagation tailored for SSM.
% Notably, the Retentive Network (RetNet)~\cite{sun2024retnet} evolves from Mamba principles to combine the parallelism of Transformers with the recurrence of RNNs, providing a promising foundation for large language models with long-range dependencies.

\subsection{Prompt-based Learning} 
Prompt learning was initially designed for adapting large-scale pre-trained language models to downstream tasks. Its core idea is to guide the model to produce the desired output by constructing prompts, rather than retraining the entire model or most of its parameters.
In the field of computer vision, prompt learning has gradually become an effective approach to improving model performance and reducing computational overhead. For instance, Jia et al.~\cite{jia2022visual} propose VPT, which introduces a small number of trainable parameters into the input space, and the backbone of the model is kept frozen, achieving efficient model fine-tuning. Zhu et al.~\cite{zhou2022learning} propose CoOp, which aims to introduce a learnable prompt approach to better adapt the powerful and generalized priors of visual language models (such as CLIP~\cite{radford2021learning}) to downstream tasks.

With the rise of prompt learning in vision tasks, recent work began to incorporate prompting mechanisms into the visual object tracking task to improve tracking accuracy. 
% Wang et al.~\cite{wang2025prior} introduce a hybrid prompt learning framework driven by prior semantic knowledge. 
SNNPTrack~\cite{ji2025snnptrack} employs a bioinspired approach, using a spiking neural network to design an event-aware prompt module, effectively improving the temporal sensitivity and energy efficiency of RGB-Eventdata. 
ViPT, proposed by Zhu et al.~\cite{zhu2023visual}, which learns modality-related cues using prompt learning for visual object tracking.
Wang et al.~\cite{wang2024temporal} propose a novel temporal adaptive RGBT tracking framework called TATrack, which achieves comprehensive utilization of spatiotemporal and multimodal information using modality-complementary prompter.
Shi et al.~\cite{shi2024explicit} generate explicit visual cues to facilitate inference in the current frame, and consider multi-scale information as explicit visual cues to achieve effective object tracking.
Different from the above works, in this paper, we design a novel Prompt Generator to generate two modality-specific learnable prompts from both RGB and event modalities, enabling a simple yet efficient multimodal fusion approach for visual object tracking.

\section{Our Proposed Approach}
\label{sec:approach}

\subsection{Overview}
\label{subsec:overview}

\begin{figure*}
\centering
\includegraphics[width=\textwidth]{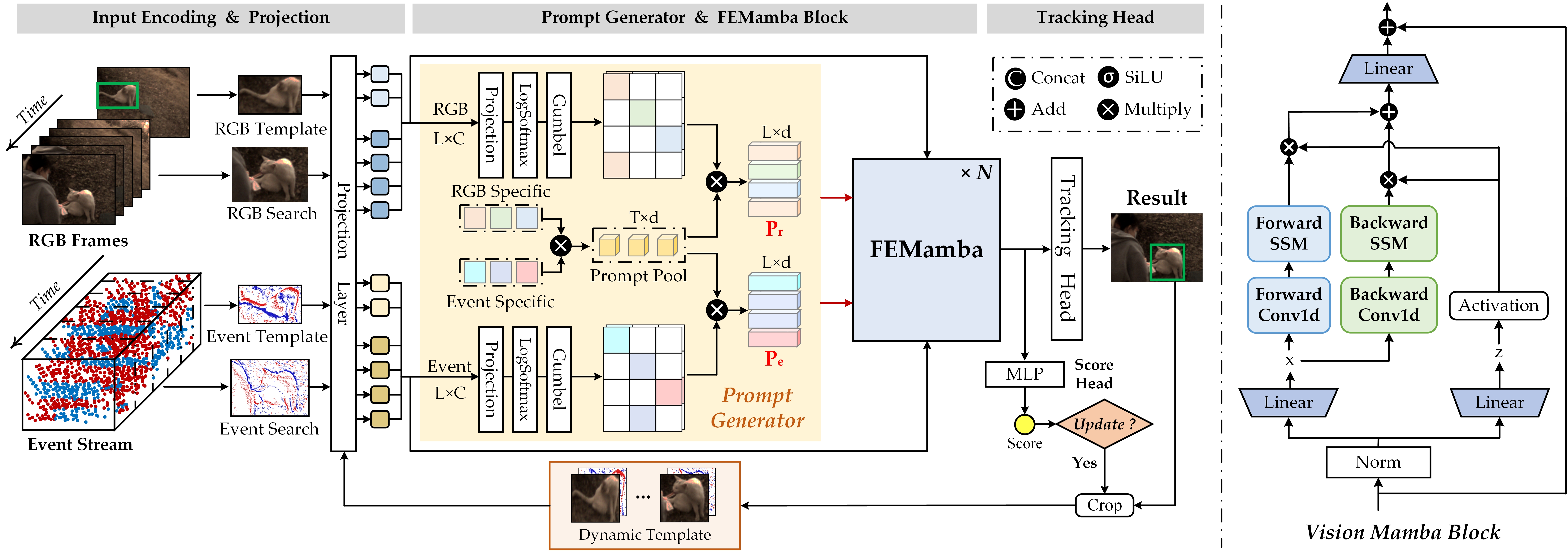}
\caption{An overview of our proposed Mamba-FETrack V2 framework. Given the RGB frames and Event streams, we first extract the template and search regions of dual modalities and transform them into tokens. Then, the newly proposed Prompt Generator is adopted to produce the modality-specific prompts dynamically. The FEMamba network is proposed to achieve feature extraction, fusion, and interaction for enhanced RGB-Event feature learning. These features are concatenated and fed into the tracking head for localization of the target object. We also introduce the dynamic template update scheme to handle the issue of appearance variation in the tracking task. 
}
\label{framework}
\end{figure*}

In this section, we will give an overview of the proposed Mamba-FETrack V2 framework, designed for visual object tracking by effectively fusing information from RGB frames and event streams. 
As illustrated in Fig.~\ref{framework}, the architecture begins by processing the input RGB template/search region and the corresponding event stream representations. These inputs are transformed into token sequences and added with position encodings. A key component of the framework is the novel Prompt Generator, which dynamically creates modality-specific prompts, denoted as \(P_r\) for RGB and \(P_e\) for event modality, based on the initial visual representations. Subsequently, a compact FEMamba network is employed to perform joint feature extraction, fusion, and interaction. The generated prompts \(P_r\) and \(P_e\) are integrated directly into the core computation of two parallel Vision Mamba blocks, effectively steering the feature extraction process and facilitating cross-modal interaction. The resulting enhanced features are concatenated and subsequently fed into the tracking head to predict the target object's location using the response maps. In addition, we also employ the dynamic template mechanism to handle the RGB-Event long-term visual tracking task.

\subsection{Input Representation}
\label{subsec:input_representation}

The Mamba-FETrack V2 framework takes as input two primary modalities: a sequence of RGB video frames, denoted as \(I = \{I_1, I_2, \dots, I_N\}\), and the corresponding asynchronous event streams from the event camera, represented as \(E = \{e_1, e_2, \dots, e_M\}\). Following standard practice in visual tracking (e.g., OSTrack \cite{ye2022joint}, Mamba-FETrack \cite{huang2024mamba}), we extract the target template $Z_I \in \mathbb{R}^{B \times 3 \times 128 \times 128}$ and the search region $X_I \in \mathbb{R}^{B \times 3 \times 256 \times 256}$ from the RGB frames.
To enable processing analogous to that of RGB frames, the sparse and asynchronous event streams are converted into a dense representation suitable for standard visual models. Specifically, we stack the event stream occurring within the exposure time corresponding to each RGB frame, generating an aligned event frame. From these event frames, we similarly extract the event template $Z_E \in \mathbb{R}^{B \times 3 \times 128 \times 128}$ and event search region $X_E \in \mathbb{R}^{B \times 3 \times 256 \times 256}$, respectively.

These four images \((Z_I, X_I, Z_E, X_E)\) are first divided into a sequence of non-overlapping small patches. Then, a projection layer (Patch Embedding) maps each small patch to a fixed-dimensional token representation. To preserve crucial spatial information, learnable absolute position encodings are added to the token sequences corresponding to the template and search regions, respectively. This process yields four initial token sequences, namely RGB template tokens \(H_I^z \in \mathbb{R}^{B \times N_z \times C}\), RGB search tokens \(H_I^x \in \mathbb{R}^{B \times N_x \times C}\), event template tokens \(H_E^z \in \mathbb{R}^{B \times N_z \times C}\), and event search tokens \(H_E^x \in \mathbb{R}^{B \times N_x \times C}\), where \(B\) is the batch size, \(C\) is the embedding dimension, and \(N_z, N_x\) are the number of tokens for template and search regions, respectively. These sequences serve as the input to the subsequent Prompt Generator and FEMamba backbone networks.

\subsection{Prompt Generator}
\label{subsec:prompt_generator}

To facilitate adaptive guidance and interaction between multi-modal features, we introduce the Prompt Generator module. Its core function is to dynamically generate modality-specific learnable prompt vectors based on the input RGB and event embeddings. This module consists of three key components: learnable prompt pools, routing-based prompt selection, and final prompt generation, as shown in Fig.~\ref{framework}.

First, we define two learnable embedding matrices as the basis for modality-specific prompts: \(W_{r} \in \mathbb{R}^{T \times d}\) and \(W_{e} \in \mathbb{R}^{T \times d}\), where \(T\) is the number of basis prompts and \(d\) is the dimension of each basis prompt. These matrices are combined through element-wise multiplication to construct a shared, learnable prompt pool \(P_{pool}\) $\in \mathbb{R}^{T \times d}$:
\begin{equation}
    P_{pool} = W_{r} \odot W_{e} .
    \label{eq:prompt_pool}
\end{equation}
This approach, empirically found to be effective, potentially introduces beneficial non-linear interactions compared to simpler methods like concatenation or addition, providing a richer semantic basis for prompt selection.

Next, a routing-based prompt selection mechanism allows the model to dynamically choose relevant prompts from the prompt pool based on the input modality feature. Concretely, the concatenated RGB tokens \(H_{rgb} = [H_I^z, H_I^x] \in \mathbb{R}^{B \times L \times C}\) and event tokens \(H_{event} = [H_E^z, H_E^x] \in \mathbb{R}^{B \times L \times C}\) (where \(L = N_z + N_x\)) are passed through a shared routing network \(F_{route}\). This routing network consists of an MLP with LogSoftmax function, and outputs log-probability distributions $R_{r}^{log} \in \mathbb{R}^{B \times L \times T}$ and $R_{e}^{log} \in \mathbb{R}^{B \times L \times T}$, which indicating the relevance of each token to each basis prompt in the pool:
\begin{align}
    R_{r}^{log} &= F_{route}(H_{rgb}), \label{eq:route_rgb} \\
    R_{e}^{log} &= F_{route}(H_{event}). \label{eq:route_evt}
\end{align}
To make the routing process differentiable, we apply the Gumbel-Softmax trick \cite{tucker2017rebar} to the log-probabilities, enabling the generation of one-hot routing matrices $O_{r} \in \mathbb{R}^{B \times L \times T}$ and $O_{e} \in \mathbb{R}^{B \times L \times T}$:
\begin{align}
    O_{r} &= {Gumbel-Softmax}(R_{r}^{log}), \label{eq:gumbel_rgb} \\
    O_{e} &= {Gumbel-Softmax}(R_{e}^{log}). \label{eq:gumbel_evt}
\end{align}
Each row of these one-hot matrices effectively identifies the prompt associated with each input token, thereby facilitating adaptive prompt selection based on the current visual representation.

Finally, in the dynamic prompt generation stage, the resulting routing matrices are matrix-multiplied with the shared prompt pool \(P_{pool}\) to produce the final dynamic prompt vectors $P_r \in \mathbb{R}^{B \times L \times d}$ and $P_e \in \mathbb{R}^{B \times L \times d}$ for the RGB and event modalities, respectively. The formula can be written as: 
\begin{align}
    P_r &= O_{r} \times P_{pool},  \label{eq:prompt_r} \\
    P_e &= O_{e} \times P_{pool}. \label{eq:prompt_e}
\end{align}

These dynamically generated prompts encapsulate informative features that are carefully tailored to the unique characteristics of each input modality. By capturing context-aware and modality-relevant representations, these prompts enhance the model’s ability to focus on salient information adaptively. Once generated, these learnable prompts are integrated into the FEMamba backbone network, where they play a crucial role in guiding the process of feature fusion and enabling deep interactive learning between two modalities.

\begin{figure*}
    \centering
    \includegraphics[width=0.9\textwidth]{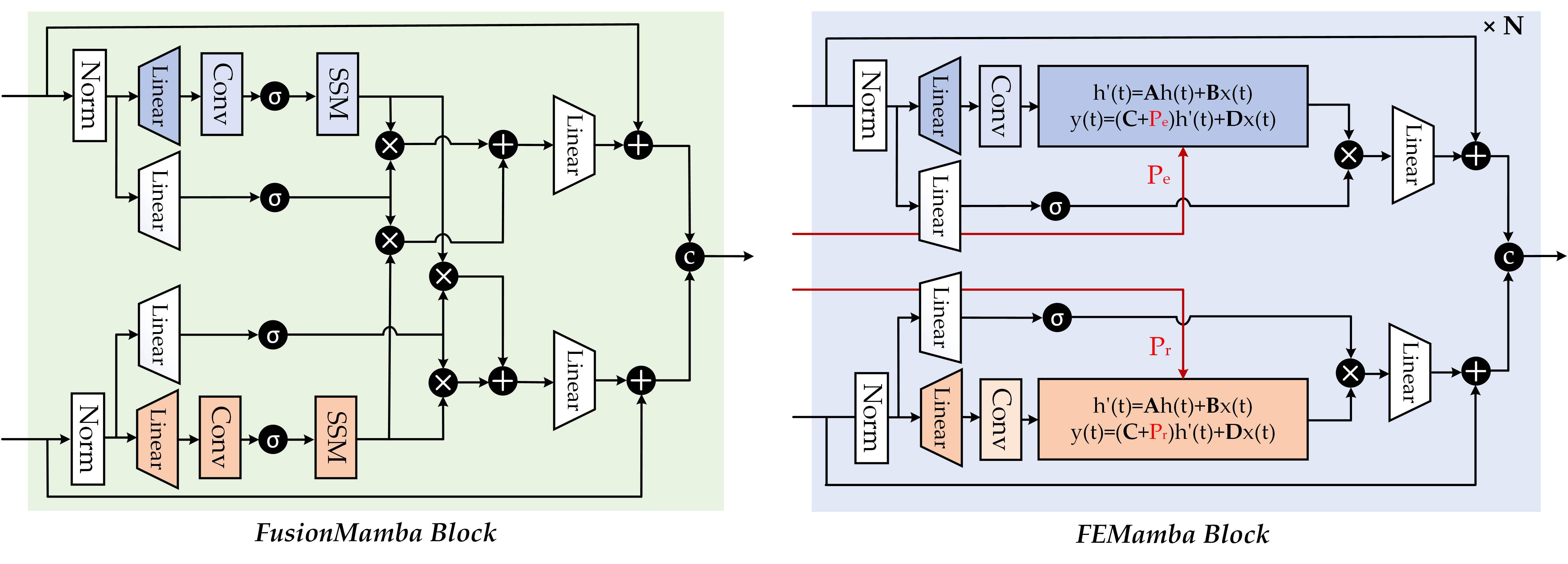}
    \caption{A comparison between FusionMamba block (proposed in our conference paper~\cite{huang2024mamba}) and the newly proposed FEMamba block. }
    \label{fig:sub_framework}
\end{figure*}

\subsection{FEMamba Backbone}
\label{subsec:fe_mamba}

To effectively extract modality-specific features and facilitate deep cross-modal interaction, we propose a unified FEMamba backbone network, which consists of two parallel Vision Mamba blocks sharing the same architecture, as shown in the right figure of Fig.~\ref{fig:sub_framework}. Crucially, the prompt integration adopts a cross-modal strategy: the RGB Vim backbone learns the RGB features \(H_{rgb}\) by incorporating the event prompt \(P_e\), whereas the event backbone extracts the event features \(H_{event}\) guided by the RGB prompt \(P_r\). In the following, we take the RGB branch as an example for clarity.

Let's detail the process within the \(l\)-th FEMamba block of the RGB branch, which receives \(H_{rgb}^{(l-1)}\) and the event prompt \(P_e\) as inputs.
First, the input tokens are normalized through the Normalization layer, and using parallel linear projections, generate intermediate representations \(z_{rgb}\) and \(x_{rgb}\). It can be mathematically represented as:
\begin{align}
    z_{rgb} &= \text{Linear}_z(\text{Norm}(H_{rgb}^{(l-1)})) \label{eq:mamba_z}, \\
    x_{rgb} &= \text{Linear}_x(\text{Norm}(H_{rgb}^{(l-1)})) \label{eq:mamba_x}.
\end{align}
The \(x_{rgb}\) representation is processed through a 1D depth-wise convolution followed by a SiLU activation to produce \(x_{rgb}'\):
\begin{equation}
    x_{rgb}' = \text{SiLU}(\text{Conv1d}(x_{rgb})).
    \label{eq:mamba_x_prime}
\end{equation}
Subsequently, \(x_{rgb}'\) is linearly projected to generate the state space matrices, \(B\), \(C\), and \(\Delta\):
\begin{equation}
    B, C, \Delta = \text{Linear}(x_{rgb}').
    \label{eq:mamba_params}
\end{equation}
Using \(\Delta\), the discrete state matrix \(\bar{A}\) and input matrix \(\bar{B}\) are obtained via the Zero-Order Hold (ZOH) discretization rule applied to the base state space matrices \(A\) and \(B\):
\begin{equation}
    \bar{A}, \bar{B} = \text{ZOH}(A, B, \Delta).
    \label{eq:mamba_discretize}
\end{equation}

Here lies the core prompt-guided cross-modal fusion (PCMF) step. Prior to executing the selective scan, the dynamic event prompt \(P_e\) generated based on event features is added to the original output matrix \(C\) in the RGB branch. This creates a cross-modality enhanced output matrix, avoiding redundant networks for interactive learning. Therefore, the formula of SSM in the RGB branch can be denoted as:
\begin{align}
   h'(t) &= \bar{A} h(t) + \bar{B} x(t), \\
   y(t) &= (C+P_e) h'(t) + D x(t),
   \label{eq:prompt_integrate_rgb_uses_event}
\end{align}
where $h(t)$ and $h'(t)$ denote the status at the previous moment and the current moment, respectively. $x(t)$ and $y(t)$ are the input and output at the current moment.
Through this mechanism, the event modality features are adaptively transferred to the RGB modality via the event prompt $P_e$, facilitating natural and efficient cross-modal feature interaction learning from event to RGB domains. Integrating the prompt by modifying \(C\) is advantageous due to its ease of implementation and validation, negligible computational and memory overhead, and its avoidance of potential gradient issues (e.g., explosion or vanishing) often associated with altering the core SSM state transition dynamics (\(\bar{A}, \bar{B}\)), thus enabling training stability.

Through selective scanning in both forward and backward directions, we will obtain $y_f$ and $y_b$, respectively. These outputs are subsequently gated by $z$ and summed:
\begin{equation}
    y_{rgb} = y_f \odot \text{SiLU}(z_{rgb}) + y_b \odot \text{SiLU}(z_{rgb}),
    \label{eq:mamba_gate}
\end{equation}
where SiLU (Sigmoid-weighted Linear Unit) refers to the activation function.
Finally, a residual connection yields the block's output. The formula is as follows:
\begin{equation}
    H_{rgb}^l = H_{rgb}^{(l-1)} + \text{Linear}(y_{rgb}),
    \label{eq:mamba_residual}
\end{equation}
where the $H_{rgb}^l$ represents the output of the RGB branch at the l-th layer.

The event branch adopts a symmetric architecture to maintain modality balance. As input, it receives the event token sequence from the previous layer \(H_{event}^{(l-1)}\) along with the RGB prompt \(P_r\). Subsequently, a parallel Vision Mamba block implements a prompt fusion mechanism, integrating the RGB prompt \(P_r\) into the event modality through \(C' = C + P_r\). This operation enables cross-modal guidance from the RGB prompt while preserving the structural integrity of event representations, thereby allowing the distilled RGB semantics to adaptively steer event feature extraction. 

After stacking \(N\) such FEMamba blocks with this cross-modal prompt injection, we obtain the final enhanced features \(F_{rgb} = H_{rgb}^L\) and \(F_{event} = H_{event}^L\). These features encapsulate rich, modality-specific information that has been mutually enhanced through deeply integrated cross-modal guidance before being forwarded to the tracking head. 

To better accommodate the continuous variations in the target's appearance, we empirically integrate a template updating strategy inspired by STARK~\cite{yan2021learning}. As illustrated in Fig.~\ref{framework}, the framework maintains both an initial static template and an adaptively updated dynamic template. Specifically, the features extracted by the FEMamba backbone are fed into a score prediction head based on an MLP architecture, which estimates the reliability of the current frame. When a predefined update interval is reached and the confidence score exceeds a threshold of 0.5, the current frame is cropped based on the tracking result, and the dynamic template is subsequently updated. This strategy enables timely adaptation to appearance variations, effectively mitigating performance degradation caused by challenges such as target deformation.

\subsection{Tracking Head and Loss Function}
\label{subsec:head_loss}

Our tracking head design closely follows OSTrack \cite{ye2022joint}. The enhanced search region features extracted from the FEMamba backbones, encompassing both RGB and event modalities, are concatenated along the channel dimension. This multimodal feature representation is subsequently processed by the tracking head to estimate the target object's spatial position.

Specifically, the concatenated features are first reshaped into a feature map format. This map is then processed by a stack of Convolution-Batch Normalization-ReLU (Conv-BN-ReLU) layers. The head outputs four main components: a target classification score map, predicting the likelihood of the target presence at each location; local offsets for refining the bounding box center coordinates; normalized bounding box dimensions (width and height); and the predicted bounding box, which denote the predicted location for the target in the current frame.

During training, we first adopt a loss structure similar to OSTrack, employing three distinct loss functions for comprehensive optimization: Focal Loss (\(L_{focal}\)) for classification, L1 Loss (\(L_1\)) for offset regression, and GIoU Loss (\(L_{GIoU}\)) for bounding box size and overlap regression. The total loss function \(L_{total}\) is a weighted sum of these components:
\begin{equation}
    L_{total} = \lambda_1 L_{focal} + \lambda_2 L_1 + \lambda_3 L_{GIoU}
    \label{eq:total_loss}
\end{equation}
where \(\lambda_1, \lambda_2, \lambda_3\) are weighting coefficients balancing the contribution of each loss term, which are set to 1, 14, and 1, respectively. After training the base model, we follow the approach of STARK~\cite{yan2021learning} by incorporating a template updating module to further train the score prediction head. We employ the commonly used Binary Cross-Entropy with Logits Loss (BCEWithLogitsLoss) to supervise the score head. The formulation is given as follows:
\begin{equation}
\mathcal{L}_{\text{bce}} = -\frac{1}{N} \sum_{i=1}^{N} \left[ y_i \cdot \log \sigma(x_i) + (1 - y_i) \cdot \log (1 - \sigma(x_i)) \right],
\end{equation}
where $x_i$ and $y_i$ denote the predicted logit and the corresponding ground-truth label, respectively, and $\sigma(\cdot)$ represents the sigmoid activation function. These loss functions jointly guide the network to learn accurate target localization and scale estimation.

\section{Experiments} \label{sec::Experiments}

\subsection{Datasets and Evaluation Metric} 
To comprehensively evaluate the effectiveness of the proposed method, we conducted extensive experiments on three benchmark datasets: FE108, FELT V2, and COESOT. These datasets cover a wide range of challenging scenarios in RGB-Event visual tracking. A detailed description of each dataset is provided below to illustrate its characteristics and the specific challenges.

\noindent $\bullet$ \textbf{FE108 Dataset}: This dataset is a dual-modality single object tracking dataset collected using a grayscale DAVIS 346 event camera. It consists of 108 tracking videos captured in indoor scenes, with a total duration of approximately 1.5 hours. Among them, 76 videos are used for training and 32 for testing. The dataset covers 21 different object categories, which can be grouped into three main types: animals, vehicles, and daily objects. In addition, it includes four challenging scenario types: low-light (LL), high dynamic range (HDR), fast motion with and without motion blur on APS frame (FWB and FNB). Please refer to the following GitHub for more details \url{https://github.com/Jee-King/ICCV2021_Event_Frame_Tracking?tab=readme-ov-file}.

\noindent $\bullet$ \textbf{FELT V2 Dataset}: This dataset is a large-scale, long-term, dual-modality single-object tracking benchmark, captured using the DAVIS 346 event camera. It contains a total of 1044 video sequences, with 730 sequences used for training and 314 for testing. Each sequence lasts on average more than 1.5 minutes, ensuring that each video includes at least 1,000 synchronized pairs of RGB and event frames. In total, it includes 1,949,680 annotated frames, encompasses 60 object categories, and defines 14 challenging tracking attributes, such as occlusion, fast motion, and low illumination. Please refer to the following GitHub for more details \url{https://github.com/Event-AHU/FELT_SOT_Benchmark}.

\noindent $\bullet$ \textbf{COESOT Dataset}: 
It is a category-wide RGB-Event-based tracking benchmark designed to evaluate the generalization ability of tracking algorithms across diverse object types. It comprises a total of 1,354 video sequences spanning 90 object categories, with 478,721 annotated RGB frames. The dataset is systematically divided into 827 training videos and 527 testing videos. To facilitate fine-grained performance analysis, 17 challenging factors—including fast motion, occlusion, illumination variation, background clutter, and scale change—are explicitly defined. Please refer to the following GitHub for more details \url{https://github.com/Event-AHU/COESOT}.

% \noindent $\bullet$ \textbf{VisEvent Dataset}: The VisEvent dataset is the first large-scale frame-event tracking dataset recorded using a color DVS346 event camera, offering high temporal resolution along with synchronized RGB and event data. It consists of 820 video sequences captured across various indoor and outdoor environments, ensuring diverse scene coverage and motion patterns. For experimental consistency, the dataset is divided into 500 training videos and 320 testing videos. More details can be found on GitHub \url{https://github.com/wangxiao5791509/VisEvent_SOT_Benchmark}.

For evaluation metrics, we adopt three widely used metrics in tracking tasks: \textbf{Precision (PR)}, \textbf{Normalized Precision (NPR)}, and \textbf{Success Rate (SR)}. Specifically, Precision (PR) represents the proportion of frames where the error between the predicted center position and the ground truth center position is smaller than a predefined threshold (default is 20 pixels). Normalized Precision (NPR) is calculated by measuring the Euclidean distance between the predicted center and the ground truth center, and then scaling it using the diagonal matrix formed by the width and height of the ground truth bounding box for scale normalization. Success Rate (SR) represents the percentage of frames where the Intersection over Union (IoU) between the predicted bounding box and the ground truth bounding box exceeds a given threshold.

\subsection{Implementation Details} 
Our Mamba-FETrack V2 framework is built upon the OSTrack architecture~\cite{ye2022joint}, with a key modification of replacing the original vision Transformer backbone with the more efficient Vision Mamba network~\cite{vim}, which significantly enhances the model’s efficiency while maintaining competitive accuracy. The model is trained using the AdamW optimizer\cite{loshchilov2018adamw}, with a learning rate set to 0.0001, a weight decay of 0.0001, and a batch size of 48. To detect whether the target has drifted outside the search region, we monitor the maximum response score in the current frame. If the maximum scores remain below 0.3 for k consecutive frames (with k=8 on FE108 and FELT V2, and k=2 on COESOT), we enlarge the cropping ratio of the search region by a factor of 1.5. This adaptive adjustment helps the tracker to re-locate fast-moving or abruptly shifting targets that may have left the original search region. All experiments are implemented using the PyTorch framework~\cite{paszke2019pytorch} and conducted on a computing server equipped with an AMD EPYC 7542 32-core CPU and an NVIDIA RTX 4090 GPU. Further implementation details are available in our source code.

\subsection{Comparison on Public Benchmark Datasets}

\begin{table*}
\center
\small     
\caption{Experimental results (SR/PR) on FE108 dataset.} 
\label{FE108table}
% \resizebox{\columnwidth}{!}{
\begin{tabular}{c|cccccc}
\hline  \toprule [0.5 pt] 
\textbf{Tracker}  &\textbf{SiamRPN}~\cite{Li2018HighPV}   &\textbf{SiamFC++}~\cite{xu2020siamfc++}   &\textbf{KYS}~\cite{bhat2020know}  &\textbf{CLNet}~\cite{dong2020clnet}  &\textbf{CMT-MDNet}~\cite{Wang2021VisEventRO}  &\textbf{ATOM}~\cite{Danelljan2018ATOMAT}   \\
\hline
\textbf{SR/PR}  &21.8/33.5  &23.8/39.1 &26.6/41.0 &34.4/55.5 &35.1/57.8  &46.5/71.3 \\
\hline
\textbf{Tracker}   &\textbf{DiMP}~\cite{Bhat2019LearningDM}     &\textbf{PrDiMP}~\cite{Danelljan2020ProbabilisticRF}     &\textbf{CMT-ATOM}~\cite{Wang2021VisEventRO}  &\textbf{CEUTrack}~\cite{tang2022coesot}    &\textbf{Mamba-FETrack}~\cite{huang2024mamba}  &\textbf{Ours}\\
\hline
\textbf{SR/PR} &52.6/79.1 &53.0/80.5 &54.3/79.4 &55.58/84.46  & 58.71/90.95 &62.18/94.74\\

\hline  \toprule [0.5 pt]  
\end{tabular}
% }
\end{table*}	

\noindent $\bullet$ \textbf{Results on FE108~}
Table~\ref{FE108table} presents a performance comparison between our proposed method and several state-of-the-art (SOTA) approaches on the FE108 dataset. The results demonstrate that our method achieves the top rank among all compared methods, with significant improvements of 94.74 on PR and 62.18 on SR. Compared to Mamba-FETrack~\cite{huang2024mamba}, the proposed Mamba-FETrack V2 shows consistent performance gains in both PR and SR metrics, highlighting the effectiveness of our feature fusion strategy.

\begin{figure*}[!htp]
\centering
\includegraphics[width=\textwidth]{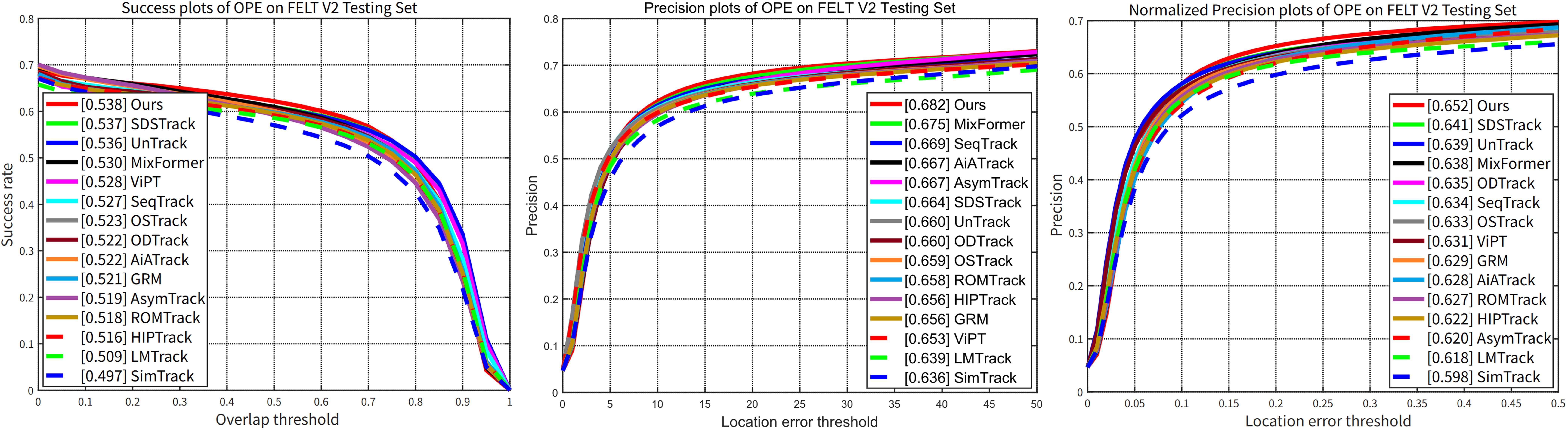}
\caption{Visualization of tracking results on the FELT V2 dataset.}  
\label{SRPRNPR}
\end{figure*}

\noindent $\bullet$ \textbf{Results on FELT V2~} 
As shown in Table~\ref{FELTV2_benchmark_results}, we compare the experimental results of our method with other SOTA tracking approaches on the FELT V2 dataset. Specifically, the classic ViT-based tracking algorithm OSTrack achieves a 52.3, 65.9, and 63.3 on SR, PR, and NPR, respectively. In comparison, our method surpasses OSTrack, achieving 53.8, 68.2, and 65.2 on the same three evaluation metrics, respectively. Furthermore, compared to other competitive trackers such as ViPT, our method demonstrates a superior performance, achieving a 2.9 percentage point improvement in precision. These results clearly highlight the effectiveness of our approach on this challenging long-term tracking benchmark. Fig.~\ref{SRPRNPR} visualizes the performance comparison with other SOTA methods on the FELT V2 dataset.

\begin{table}
\centering
\small
\caption{Tracking results on FELT V2 Dataset.} 
\label{FELTV2_benchmark_results}
% \resizebox{\textwidth}{!}{ 
\begin{tabular}{l|l|c}
\hline \toprule [0.5 pt]
\textbf{Trackers} & \textbf{Source}  &\textbf{SR/PR/NPR} \\
\hline 
% &   \textbf{TransT~\cite{Chen2021TransformerT}} & CVPR21  &   &   \\
% &   \textbf{STRAK~\cite{Yan2021LearningST}} & ICCV21  & & \\
\textbf{01. OSTrack~\cite{ye2022joint}} & ECCV22  & 52.3/65.9/63.3  \\
\textbf{02. MixFormer~\cite{Cui2022MixFormerET}} & CVPR22  & 53.0/67.5/63.8 \\
   \textbf{03. AiATrack~\cite{gao2022aiatrack}} & ECCV22 & 52.2/66.7/62.8   \\
   \textbf{04. SimTrack~\cite{chen2022backbone}} & ECCV22  & 49.7/63.6/59.8   \\
   \textbf{05. GRM~\cite{Gao2023GeneralizedRM}} & CVPR23  & 52.1/65.6/62.9    \\
   \textbf{06. ROMTrack~\cite{cai2023robust}} & ICCV23  & 51.8/65.8/62.7   \\
% &   \textbf{CiteTrack~\cite{li2023citetracker}} & ICCV23 &  &  \\
   \textbf{07. ViPT~\cite{zhu2023visual}} & CVPR23 &52.8/65.3/63.1  \\
   \textbf{08. SeqTrack~\cite{chen2023seqtrack}} & CVPR23 & 52.7/66.9/63.4   \\
% \textbf{SeqTrackv2~\cite{chen2024unifiedsequencetosequencelearningsingle}} & CVPR23 & 55.5/68.6/65.6 & \\
% &   \textbf{ARTrackv2~\cite{bai2024artrackv2}} & CVPR24  &   & \\
   \textbf{09. HIPTrack~\cite{cai2024hiptrack}} & CVPR24  & 51.6/65.6/62.2  \\
   \textbf{10. ODTrack~\cite{Zheng2024ODTrackOD}} & AAAI24  &52.2/66.0/63.5   \\
% \textbf{EVPTrack~\cite{shi2024explicit}} & AAAI24 &53.8/68.7/64.8  & \\
% \textbf{AQATrack~\cite{xie2024autoregressive}} & CVPR24 & 54.0/69.1/64.7  & \\
   \textbf{11. SDSTrack~\cite{hou2024sdstrack}} & CVPR24  & 53.7/66.4/64.1    \\
   \textbf{12. UnTrack~\cite{wu2024single}} & CVPR24 & 53.6/66.0/63.9  \\
% &   \textbf{13. SUTrack~\cite{chen2025sutrack}} & AAAI25  & 56.6/70.9/66.6  \\
 \textbf{13. LMTrack~\cite{xu2025less}} & AAAI25 & 50.9/63.9/61.8  \\
   \textbf{14. AsymTrack~\cite{zhu2025two}} & AAAI25 & 51.9/66.7/62.0  \\
\hline
   \textbf{15. Ours} &- &53.8/68.2/65.2   \\
\hline \toprule [0.5 pt]
\end{tabular} 
% } 
\end{table}

\noindent $\bullet$ \textbf{Results on COESOT~} 
As shown in Table~\ref{coesot_result}, we also report our tracking results on the large-scale RGB-Event tracking dataset COESOT. Compared to most popular tracking algorithms, such as OSTrack and AiATrack, our method demonstrates significant advantages across multiple metrics, achieving SR and PR scores of 62.6 and 76.9, respectively. These results highlight the superior tracking performance of our method across various target categories in the dataset.
However, when compared to several robust methods built on high-complexity ViT architectures, such as ViPT, our method shows a disadvantage in terms of SR. This suggests that there is still room for further improvement in bounding box alignment accuracy.

\begin{table}
\centering
\small
\caption{Tracking results on COESOT Dataset.} 
\label{coesot_result}
% \resizebox{\textwidth}{!}{ 
\begin{tabular}{l|l|cc}
\hline \toprule [0.5 pt]
\textbf{Trackers} &\textbf{Source}  &\textbf{SR}  &\textbf{PR} \\
\hline 
\textbf{01. TransT~\cite{Chen2021TransformerT}} & CVPR21  &60.5   &72.4   \\
   \textbf{02. STRAK~\cite{Yan2021LearningST}} & ICCV21  &56.0 &67.7 \\
   \textbf{03. OSTrack~\cite{ye2022joint}} & ECCV22  &59.0  &70.7 \\
   \textbf{04. MixFormer~\cite{Cui2022MixFormerET}} & CVPR22  &55.7 &66.3 \\
   \textbf{05. AiATrack~\cite{gao2022aiatrack}} & ECCV22 &59.0 &72.4  \\
   \textbf{06. SiamR-CNN~\cite{voigtlaender2020siam}} & CVPR20  &60.9  &71.0     \\
   \textbf{07. ToMP50~\cite{Mayer2022TransformingMP}} & CVPR22  &59.8 &70.8 \\
   \textbf{08. ToMP101~\cite{Mayer2022TransformingMP}} & CVPR22 &59.9 &71.6 \\
   \textbf{09. KeepTrack~\cite{mayer2021learning}} & ICCV21  &59.6  &70.9  \\
% &   \textbf{CEUTrack~\cite{tang2022coesot}} & arXiv22  & &  \\
% &   \textbf{CEUTrack+~\cite{tang2022coesot}} & arXiv22  &62.7 &76.0  \\
   \textbf{10. PrDiMP50~\cite{danelljan2020probabilistic}} & CVPR20  &57.9 &69.6  \\
% &   \textbf{11. KYS~\cite{bhat2020know}} &ECCV20   &58.6  &71.6  \\
   \textbf{11. DiMP50~\cite{bhat2019learning}} &ICCV19  &58.9 &72.0 \\
   \textbf{12. ATOM~\cite{danelljan2019atom}} &CVPR19   &55.0 &68.8  \\
   \textbf{13. TrDiMP~\cite{wang2021transformer}} &CVPR21  &60.1 &72.2  \\
   \textbf{14. MDNet~\cite{Wang2021VisEventRO}} &TCYB23   &53.3 &66.5 \\
   \textbf{15. ViPT~\cite{zhu2023visual}} &CVPR23   &66.3 &74.5 \\
% &   \textbf{15. HRCEUTrack~\cite{zhu2023cross}} &ICCV23  &63.2  &71.9  \\
   \textbf{16. CMDTrack~\cite{zhang2025cross}} &TPAMI25   &65.7 &74.8 \\
% &   \textbf{18. MCITrack~\cite{kang2025exploring}} &AAAI25   &64.7 &78.1   \\
   \textbf{17. LMTrack~\cite{xu2025less}}   &AAAI25 & 58.4 & 71.1   \\
\hline
   \textbf{18. Ours} &- &62.6 &76.9 \\
\hline \toprule [0.5 pt]
\end{tabular} 
% } 
\end{table}

% \begin{table}[H]
% \centering
% \small     
% \caption{Experimental Results (SR/PR) on COESOT Dataset.} 
% \label{COESOT_results}
% \resizebox{\columnwidth}{!}{
% \begin{tabular}{ccccccccccc}
% \hline \toprule [0.5 pt]
% \textbf{TrDiMP~\cite{wang2021TrDiMP}}  &\textbf{ToMP50~\cite{mayer2022Tomp}}  &\textbf{OSTrack~\cite{ye2022Ostrack}}    &\textbf{AiATrack~\cite{gao2022AIa}}  &\textbf{STARK~\cite{yan2021Stark}}    &\textbf{TransT~\cite{chen2021transt}}  &\textbf{SimTrack~\cite{chen2022SimTrack}}   \\ 
% 50.7/59.2      &46.3/55.2       &50.9/61.8       &50.6/59.5        &40.8/44.5       &45.6/54.3    &48.3/55.7  \\ 
% \hline 
% \textbf{PrDiMP~\cite{martin2020PrDimp}}   &\textbf{ATOM~\cite{danelljan2019atom}}  &\textbf{MixFormer~\cite{cui2022mixformer}} &\textbf{ODTrack~\cite{zheng2024odtrack}}  &\textbf{HDETrack~\cite{wang2024event}} &\textbf{Ours-Fast}   &\textbf{Ours-Slow}     \\ 
% 47.5/57.8       &42.1/50.4          & 44.4/50.2         &51.5/62.1         &52.3/63.0   &49.3/59.1   &51.8/62.9  \\         
% \hline \toprule [0.5 pt]
% \end{tabular}
% }
% \end{table}

% \noindent $\bullet$ \textbf{Results on VisEvent} 

\subsection{Ablation Study}
To comprehensively validate the effectiveness of each component in our proposed method, we conducted comprehensive ablation studies on the FELT V2 dataset. Our analysis focuses on six key aspects: component analysis, input data, backbone architecture, fusion strategy, fusion position, and the Prompt Generator. The results of each ablation experiment are presented and discussed below.

\noindent $\bullet$ \textbf{Component Analysis.} 
As shown in Table~\ref{CAResults}, we progressively enhance the baseline model by incorporating the Prompt Generator and the Cross-modal Fusion strategy to assess their individual and combined contributions to performance. The baseline, which simply fuses RGB and event features via concatenation, achieves an SR of 50.8 and a PR of 64.7. Introducing the Prompt Generator alone improves the SR to 52.4 and the PR to 66.5. Similarly, integrating the prompt-guided Cross-modal Fusion (PCMF) module yields an SR of 52.2 and a PR of 66.1. When both modules are applied together, the model achieves an SR of 53.3 and a PR of 67.6, demonstrating the complementary advantages of the two components in enhancing multi-modal representation learning. Finally, we incorporate a template updating strategy that leverages dynamic templates to better capture the target's evolving appearance, resulting in further performance improvements.

\begin{table} 
\center
% \small   
\caption{Component Analysis on the FELT v2 Dataset. PCMF and Dyn. Temp. is short for Prompt-guided Cross-modal Fusion and Dynamic Template.} 
\label{CAResults} 
\resizebox{\columnwidth}{!}{ 
\begin{tabular}{cccc|cc} 		%% \xmark   \cmark  
\hline \toprule 
\textbf{Base}  &\textbf{Prompt Generator} &\textbf{PCMF}   &\textbf{Dyn. Temp.} &\textbf{SR} &\textbf{PR}\\
\hline 
\cmark   &          &        &       &50.8              &64.7         \\
\cmark   &\cmark    &        &       &52.4              &66.5         \\
\cmark   &          &\cmark  &       &52.2              &66.1            \\
\cmark   &\cmark    &\cmark  &       &53.3              &67.6            \\
\cmark   &\cmark    &\cmark  &\cmark &\textbf{53.8}     &\textbf{68.2}         \\
\hline \toprule  
\end{tabular}
}
\end{table}

\noindent $\bullet$ \textbf{Analysis on Input Data.~} 
In the input modality analysis (top part of Table~\ref{Ablation_Studies}), we evaluate the tracking performance using only event frames, only RGB frames, and a fusion of both modalities. When using event frames alone, the performance is relatively low, with an SR of 46.5, PR of 59.6, and NPR of 56.6. In contrast, using only RGB frames yields better results, achieving an SR of 51.8, PR of 65.6, and NPR of 63.1. When both RGB and event frames are fused, the performance is further enhanced, reaching an SR of 53.8, PR of 68.2, and NPR of 65.2. These results suggest that RGB and event modalities provide complementary information in terms of semantics and motion cues, and that their fusion significantly improves the robustness and accuracy of object tracking. 

\noindent $\bullet$ \textbf{Analysis on Backbone Network.~} 
We further compare the performance of three backbone networks: ViT-B, ViT-S, and Vim-S. As shown in Table~\ref{Ablation_Studies}, ViT-B demonstrates solid performance, with an SR of 52.3, PR of 65.9, and NPR of 63.3. However, this comes at the cost of significantly increased parameter size and computational overhead. In contrast, Vim-S achieves better performance (SR: 53.8, PR: 68.2, NPR: 65.2) while maintaining substantially lower complexity. Therefore, to better balance accuracy and efficiency, we ultimately selected Vim-S as the backbone network in our method.

\noindent $\bullet$ \textbf{Analysis on Fusion Method.~} 
We evaluate three fusion strategies: Addition, Concatenation, and Prompt-guided Cross Fusion. Among them, our proposed Prompt-guided Cross Fusion achieves the best performance with SR of 53.8, PR of 68.2, and NPR of 65.2. This demonstrates that incorporating learnable, modality-specific prompts for cross-modal interaction provides more effective guidance for fusion than simple element-wise operations. These results underscore the superiority of prompt-aware fusion in capturing cross-modal semantic relationships.

\noindent $\bullet$ \textbf{Analysis on Fusion Position.~} 
To determine the optimal fusion strategy, we further explore different fusion positions in SSM matrices: y, $\Delta$, B, and C. Among these, fusion at position C matrix achieves the best performance (SR: 53.8, PR: 68.2, NPR: 65.2), suggesting that performing fusion on the output matrix provides the greatest benefit. We believe this operation enables each modality to effectively integrate and refine high-level features received from the other modality, resulting in more effective cross-modal interaction and ultimately enhancing tracking accuracy.

\noindent $\bullet$ \textbf{Analysis on Prompt Generator.~} 
We then compared three different prompt generation strategies: Random Initialization, the absence of a Prompt Pool, and our proposed learnable Prompt Generator. As shown in Table~\ref{Ablation_Studies}, the learnable Prompt Generator consistently outperforms the other methods, demonstrating its effectiveness. This underscores the importance of modality-specific prompt learning in enhancing multi-modal fusion by providing more adaptive and informative cross-modal guidance.

% \begin{figure}[htbp]
%     \centering
%     \includegraphics[width=0.85\linewidth]{figures/FusionMethodAblation.png}
%     \caption{Fusion method ablation results.}
%     \label{fig:FusionMethodAblation}
% \end{figure}

% \begin{figure}[htbp]
%     \centering
%     \includegraphics[width=0.85\linewidth]{figures/FusionPositionAblation.png}
%     \caption{Fusion method ablation results.}
%     \label{fig:FusionPositionAblation}
% \end{figure}

% \begin{figure}[htbp]
%     \centering
%     \includegraphics[width=0.85\linewidth]{figures/PromptGeneratorAblation.png}
%     \caption{Fusion method ablation results.}
%     \label{fig:PromptGeneratorAblation}
% \end{figure}

\begin{table}
\center
\small     
\caption{Ablation Studies on the FELT V2 dataset.} 
\label{Ablation_Studies}
\resizebox{0.9\columnwidth}{!}{ 
\begin{tabular}{l|lll}
\rowcolor{gray!20}
\hline \toprule [0.5 pt] 
\textbf{\# Input Data}    &\textbf{SR}   & \textbf{PR}  & \textbf{NPR}  \\
\text{1. Event Frames only}     &46.5   &59.6   &56.6  \\
\text{2. RGB Frames only }     &51.8     & 65.6     &63.1   \\
\text{3. Event \& RGB}      & \textbf{53.8}     & \textbf{68.2}   & \textbf{65.2}   \\
\hline 
\rowcolor{gray!20}
\textbf{\# Backbone Network}    &\textbf{SR}   & \textbf{PR}  & \textbf{NPR}  \\
\text{1. ViT-B }    &52.3 &65.9 &63.3     \\ 
\text{2. ViT-S }      & 48.1    & 61.2   & 59.6      \\
\text{3. Vim-S }    & \textbf{53.8}     & \textbf{68.2}   & \textbf{65.2}    \\
\hline 
\rowcolor{gray!20}
\textbf{\# Fusion Method}    &\textbf{SR}   & \textbf{PR}  & \textbf{NPR}  \\
\text{1. Addition }                   &51.2     & 65.1      & 62.4     \\ 
\text{2. Concatenation}       &50.8    &64.7     &61.8      \\ 
\text{3. Prompt-guided Cross Fusion  }  & \textbf{53.8}     & \textbf{68.2}   & \textbf{65.2}     \\ 
\hline
\rowcolor{gray!20}
\textbf{\# Fusion Position}    &\textbf{SR}   & \textbf{PR}  & \textbf{NPR}  \\
\text{1. y }                   & 52.6    & 66.9     & 63.9      \\ 
\text{2. $\Delta$ }        & 52.3     &  66.6    & 63.8     \\ 
\text{3. B}        & 52.1     &66.1      &63.4      \\ 
\text{4. C}          & \textbf{53.8}     & \textbf{68.2}   & \textbf{65.2}   \\
\hline 
\rowcolor{gray!20}
\textbf{\# Prompt Generation}    &\textbf{SR}   & \textbf{PR}  & \textbf{NPR}  \\
\text{1.  Random Initialization}             & 52.2     &66.1      &63.2       \\ 
\text{2.  w/o Prompt Pool}              & 52.4    & 66.6    & 63.8     \\ 
\text{3.  Prompt Generator}      & \textbf{53.8}     & \textbf{68.2}   & \textbf{65.2}       \\ 
\hline \toprule [0.5 pt] 
\end{tabular}
}
\end{table}

% \subsection{Efficiency Analysis} 
% \begin{table}[ht]
% \centering
% \small     
% \caption{Comparison of FLOPs, parameters, and tracking speed.} 
% \label{Parameter}
% \resizebox{\columnwidth}{!}{
% \begin{tabular}{c|c|ccc}
% \toprule
% \textbf{Tracker} & \textbf{Backbone Type} & \textbf{FLOPs (G)} & \textbf{Params (M)} & \textbf{FPS} \\
% \midrule
% OSTrack-B  & ViT-B   & 1850 & 97  & 63 \\
% OSTrack-S  & ViT-S   & 1076 & 60  & 84 \\
% LMTrack  & ViT-B   & 69   & 92  & 47 \\
% Ours     & Vim-S   & 126   & 30  & -  \\
% \bottomrule
% \end{tabular}
% }
% \end{table}

\begin{figure*}[!htp]
\centering
\includegraphics[width=0.9\textwidth]{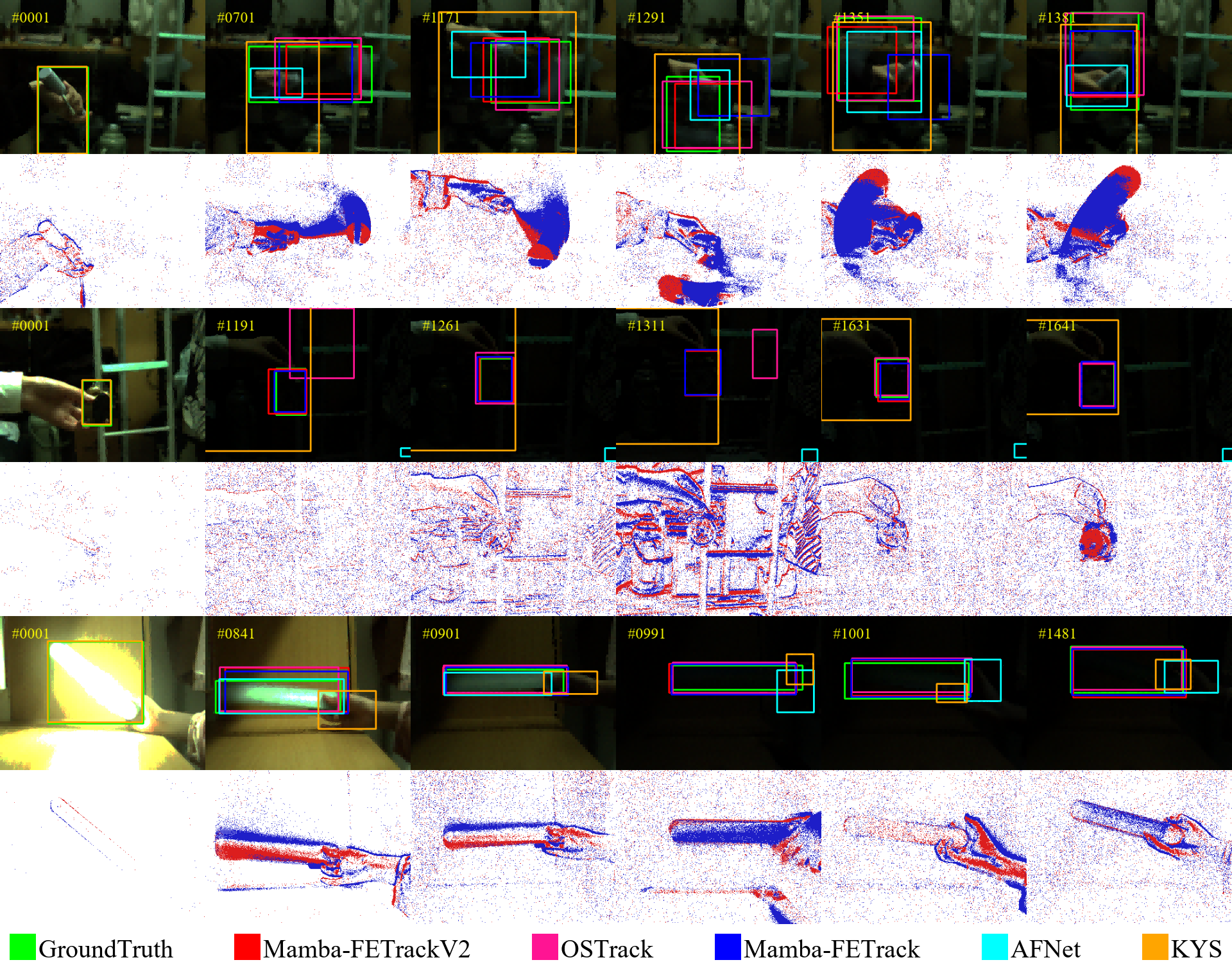} 
\caption{Qualitative comparison of tracking performance. Our Mamba-FETrack V2 (e.g., depicted with a red bounding box) consistently demonstrates superior alignment with the ground truth (e.g., green bounding box) across various challenging sequences, outperforming other representative trackers.}
\label{fig:tracking_visualization_comparison}
\end{figure*}

\subsection{Efficiency Analysis} 
To better illustrate the effectiveness and efficiency of our proposed Mamba-FETrack V2, Table~\ref{Parameter} provides a comprehensive comparison of tracking performance, model size, and inference speed on the FELT V2 dataset. Built upon the lightweight Vim-S backbone, our method achieves the highest Success Rate (SR) of 53.8, surpassing other trackers such as CEUTrack-B (52.3) and LMTrack (50.9), while maintaining the smallest model size at only 30M parameters. Compared to CEUTrack-S, which adopts a ViT-S backbone, our method delivers a significant performance improvement with approximately half the number of parameters. Additionally, our Vision Mamba-based tracker achieves an inference speed of 29 FPS. We attribute the relatively modest speed to the sequential nature of the Mamba architecture, which is inherently less parallelizable than Transformer-based models and currently lacks optimized GPU acceleration. Nevertheless, the experimental results clearly demonstrate the strength of our approach, showcasing a well-balanced trade-off between accuracy and computational efficiency.

\begin{table}
\centering
\small     
\caption{Comparison of SR, parameters, and tracking speed on the FELT V2 dataset.} 
\label{Parameter}
\resizebox{\columnwidth}{!}{
\begin{tabular}{c|c|ccc}
\toprule
\textbf{Tracker} & \textbf{Backbone} & \textbf{SR} & \textbf{Params (M)} & \textbf{FPS} \\
\midrule
CEUTrack-B  & ViT-B   & 52.3 & 97  & 28 \\
CEUTrack-S  & ViT-S   & 48.1 & 60  & 43 \\
LMTrack  & ViT-B   & 50.9   & 92  & 47 \\
Ours     & Vim-S   & 53.8   & 30  & 29  \\
\bottomrule
\end{tabular}
}
\end{table}

\begin{figure}
\centering
\includegraphics[width=0.48\textwidth]{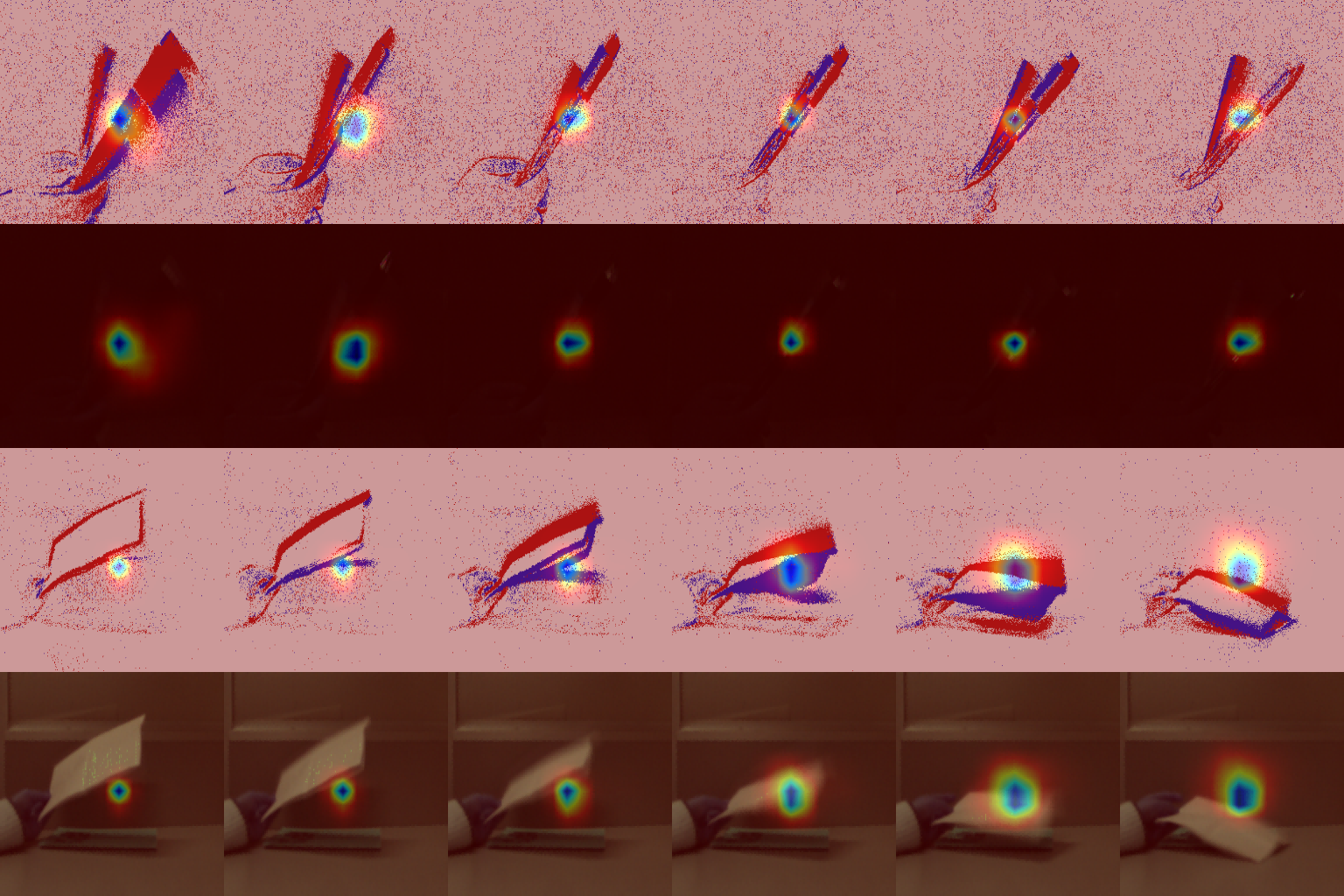}
\caption{Visualization of response maps generated by Mamba-FETrack V2.}
\label{fig:activation_maps_visualization}
\end{figure}

\subsection{Visualization}

In addition to the comprehensive quantitative analysis provided, we also present qualitative visual results to offer a more intuitive understanding of the capabilities of Mamba-FETrack V2. 
Fig.~\ref{fig:tracking_visualization_comparison} showcases a comparative visualization of our method alongside other trackers. 
It is evident that the bounding boxes predicted by Mamba-FETrack V2 maintain a closer and more consistent correspondence with the ground truth when contrasted with alternative approaches, particularly in scenarios involving complex motion or appearance changes. 
Furthermore, to illustrate the efficacy of our model in accurately focusing on the target, Fig.~\ref{fig:activation_maps_visualization} displays the response maps predicted by Mamba-FETrack V2. 
These response maps reveal that for different object categories and various common objects depicted in the sequences, our method can effectively concentrate its learned features on the designated targets, distinguishing them from the surrounding environment. 
These visualization experiments serve to intuitively illustrate the superior tracking performance and precise target localization achieved by our proposed Mamba-FETrack V2 framework.

\subsection{Limitation Analysis} 

Despite the superior accuracy and efficiency achieved by our proposed Mamba-FETrack V2, there remains room for further improvement. On one hand, designing adaptive RGB-Eventmodality fusion strategies tailored to different challenges could effectively mitigate performance degradation caused by complex tracking scenarios. On the other hand, modality imbalance during bimodal fusion may suppress the representation of the weaker modality. This issue can be alleviated by leveraging generative models to synthesize high-quality features for the weaker modality.

\section{Conclusion and Future Works} \label{sec::conclusion}

In this paper, we present Mamba-FETrack V2, a novel Frame-Event-based visual object tracking framework that introduces a new paradigm for multimodal tracking. Specifically, we first design a Prompt Generator that dynamically produces two learnable prompt vectors by leveraging both modality-specific features and a shared prompt pool. Subsequently, the generated modality-specific prompt vectors and image embedding features are jointly processed by the proposed FEMamba module, which is built upon Vision Mamba, to facilitate prompt-guided cross-modal interaction and fusion. As a result, our algorithm achieves an optimal balance between tracking accuracy and computational efficiency, establishing an efficient and high-performance solution for RGB-Event object tracking. 
In our future work, we plan to investigate more lightweight visual Mamba architectures to further enhance tracking efficiency. We believe this study will contribute to advancing Mamba-based object tracking research and inspire new developments in the field.

% \section*{Acknowledgment} \label{sec::Acknowledgment}

\small{ 
\bibliographystyle{IEEEtran}
\bibliography{reference}
}

% that's all folks
\end{document}